\documentclass[pdflatex,sn-mathphys]{sn-jnl}% Math and Physical Sciences Reference Style
\jyear{2021}%

\theoremstyle{thmstyleone}%
\theoremstyle{thmstyletwo}%

\theoremstyle{thmstylethree}%

\begin{document}

%\title[Style Based and Event Based Fashion Recommendation]{Style Based and Event Based Fashion Recommendation}
\title[Fashion Recommendation Based on Style and Social Events]{Fashion Recommendation Based on Style and Social Events}


\author*{\fnm{Federico} \sur{Becattini}}\email{federico.becattini@unifi.it}

\author{\fnm{Lavinia} \sur{De Divitiis}}\email{lavinia.dedivitiis@unifi.it}

\author{\fnm{Claudio} \sur{Baecchi}}\email{claudio.baecchi@unifi.it}

\author{\fnm{Alberto} \sur{Del Bimbo}}\email{alberto.delbimbo@unifi.it}

\affil*[1]{\orgdiv{University of Florence}, \orgname{Media Integration and Communication Center (MICC)}, \orgaddress{\street{Viale Morgagni 65}, \city{Florence}, \country{Italy}}}

% \affil[2]{\orgdiv{Department}, \orgname{Organization}, \orgaddress{\street{Street}, \city{City}, \postcode{10587}, \state{State}, \country{Country}}}

% \affil[3]{\orgdiv{Department}, \orgname{Organization}, \orgaddress{\street{Street}, \city{City}, \postcode{610101}, \state{State}, \country{Country}}}


\abstract{Fashion recommendation is often declined as the task of finding complementary items given a query garment or retrieving outfits that are suitable for a given user. In this work we address the problem by adding an additional semantic layer based on the style of the proposed dressing. We model style according to two important aspects: the mood and the emotion concealed behind color combination patterns and the appropriateness of the retrieved garments for a given type of social event. To address the former we rely on Shigenobu Kobayashi's color image scale, which associated emotional patterns and moods to color triples. The latter instead is analyzed by extracting garments from images of social events. Overall, we integrate in a state of the art garment recommendation framework a style classifier and an event classifier in order to condition recommendation on a given query.}




\maketitle

\section{Introduction}

Fashion is a way to express personal style and to communicate emotional states that reflect personality or mood. At the same time, certain social events follow a dress code which can be strictly required or implicitly followed by participants. The rules of such social requirements however are not written and can be hard to comply with, especially for first-time attendants.
Even formal events, where dressing options are more constrained, might leave space for attendants to decline an outfit according to specific preferences or personalities. Even at ceremonies such as weddings, one can adhere to the most formal requirements or play with details to give a casual or sporty touch the outfit and thus to his or her appearance.
Such customization of an outfit is a mean to communicate to others a certain intent, to establish a friendly or formal environment and even to be provocative or remark a social status.
Such moods and tastes are conveyed by both the shape and color of any fashion garment and may be subjected to trends and personal interpretations, yet color combinations and outfit making appear to be grounded on general clothing rules that are well established within the society.

%Individuals have several ways of expressing moods, emotions and feelings. Fashion for instance is a mean to communicate to others a certain intent, to establish a friendly or formal environment and even to be provocative or remark a social status.
%Such moods and tastes are conveyed by both the shape and color of any fashion garment and may be subjected to trends and personal interpretations.
% However, general clothing rules appear to be well established and grounded into how people live within the society. For instance, formal events will require a formal dressing as well as schools and offices might require a clothing conduct. Such rules mostly reflect on which category of garments an individual will wear to certain occasions and do not consider any additional feeling that one may want to communicate. For this reason, people are likely to play with color combinations to express an emotional statement or to stand out.
% \todo{estendere su eventi}

Patterns behind color-combinations have been extensively studied, in particular with reference to different styles one may want to communicate. The seminal work by Shigenobu Kobayashi~\cite{kobayashi2009color} introduced a scale to express emotions or attitudes based on color combinations.
According to Kobayashi, just considering triplets of colors, one can identify a wide variety of lifestyles which can then be expressed by personal spaces such as interiors or offices or personal items such as outfits.
On the other hand, deriving precise rules for specific events is harder since it can vary in time and follow different trends in different communities of people.

In this work we propose to follow a data driven approach to learn to model these implicit rules, both following color-combination styles and social event requirements.
For color-combinations we rely on the Kobayashi's color image scale, whereas for social events we extract from online available data information about garment appearance and the respective social event.
We intend to develop a complete recommendation system capable of taking into account the conveyed emotion and the compliance to societal standards for different types of events. We believe that this can come to the aid of people that seek help in shop assistants, either physically in shops or virtually on an online marketplace.

Since different modalities can convey different emotions or be suitable for different events, we also analyze the capabilities of a recommendation system to take into account also diversity. In general, diversity is an important factor to consider while tackling any information retrieval task. As suggested in \cite{wang2006}, this also applies to benchmark datasets and the way they are exploited to produce any recommendation. The introduction of stylish aspects based on visual cues such as color combinations is also a step towards this direction.

%In this paper we explore the integration of such emotions into a garment recommendation system, which we condition to generate outfits that satisfy a certain mood or style, following Kobayashi's color scale.

A preliminary version of our approach was described in \cite{de2021style}. The system presented in this work differs substantially
from \cite{de2021style} in several ways:
(i) instead of focusing only on Kobayashi's color image scale, we present a novel analysis concerning social events by gathering the \textit{Fashion4Events} dataset comprising approximately 400k garment images with social event labels;
(ii) using the collected dataset, we train an outfit based event classifier;
(iii) we integrate both the color classifier and the event classifier in our recommendation system in order to provide a more fine-grained filtering of the recommendations.
We also provide an improved analysis of the state of the art.

The paper is organized as follows. In Sec.~\ref{sec:related} we first provide an overview on the state of the art for fashion recommendation. On overview of our presented method is presented in Sec.~\ref{sec:overview}. Here we introduce: (i) an outfit emotion classifier based on color combinations, capable of mapping a generic outfit onto Kobayashi's color scale; (ii) a garment based social event classifier, used to infer the event category suitable for a given outfit; (iii) the integration of such modules with a state of the art garment recommendation system. The three aspects are then detailed respectively in Sec.~\ref{sec:method_style}, Sec.~\ref{sec:method_event} and Sec.~\ref{sec:method_rec}. The results of our experiments are reported in Sec.~\ref{sec:exp}. Finally, we draw conclusions in Sec.~\ref{sec:conclusions}

\section{Related work}
\label{sec:related}
Fashion recommendation must pursuit the goals of recommending pertinent items \cite{song2019gp,sarkar2022outfittransformer,de2022disentangling,baldrati2021conditioned}, providing a set of different dressing modalities \cite{divitiisbbb20,de2022disentangling,sa2022diversity} and possibly complying with some user query or desired criteria \cite{de2021style,hou2021learning}.

Several methods have focused on proposing garments in order to complement a given query item \cite{vasileva2018learning,song2019gp,divitiisbbb20}.
In particular, \cite{song2018neural} focused on complementary clothing matching, starting from a top garment and recommending a bottom item. The proposed method devised a compatibility modeling scheme with attentive knowledge distillation also exploiting a teacher-student network scheme.
The approach has then been improved by studying a personalized compatibility modeling, leveraging both general and subjective aesthetic preferences with a personalized compatibility modeling scheme named GP-BPR \cite{song2019gp}.
Still exploiting Bayesian Personalized Ranking (BPR), \cite{liu2019neural} used multiple autoencoder neural networks to leverage the multi-modalities of fashion items and their inter-compatibility.
Following up on this line of research, PAI-BPR \cite{sagar2020pai} proposed an attribute-wise interpretable compatibility scheme with personal preference modelling.

Top-bottom recommendation has also been addressed exploiting Memory Augmented Neural Networks (MANN) \cite{divitiisbbb20,de2022disentangling,de2021style}. This type of architecture exploits an external memory~\cite{graves2014neural, weston2014memory, marchetti2020memnet, marchetti2020multiple, marchetti2022smemo} by pairing different clothing items and training a memory writing controller to store a non-redundant subset of samples. This is then used to retrieve a ranked list of suitable bottoms to complement a given top.
External memories have also been used to store disentangled features for separate attributes \cite{de2022disentangling,hou2021learning}. In our work we rely on a memory based approach, namely GR-MANN \cite{divitiisbbb20}, building on top of it a style-based and event-based recommendation system.

A similar, yet more complex task is the one of generating a whole outfit from a given garment seed. One of the first approaches \cite{han2017learning} modeled sequences of suggestions by jointly learning a visual-semantic embedding and training a bidirectional LSTM (Bi-LSTM) model to sequentially predict the next item conditioned on previous ones.
\emph{Vasileva et al.} \cite{vasileva2018learning} instead learned pairwise embeddings to obtain separate representation for different pairs of fashion categories.
Recent approaches tend to represent an outfit as a graph, linking fashion items among themselves if they are compatible with each other \cite{cui2019dressing,cucurull2019context,yang2020learning}.

As for interpreting and diagnosing the proposed outfit compatibility and suggestions, \cite{wang2019outfit} learns type-specified pairwise similarities between items and uses the backpropagation gradients to diagnose incompatible factors.
Towards interpretable and customized fashion outfit compositions, \cite{feng2018interpretable} train a partitioned embedding network to favor interpretability of intermediate representations.

Finally, an interesting emerging topic is the one of understanding user reactions in order to integrate implicit user feedback into an iterative recommendation system, e.g. looking at body movements \cite{bigi2020automatic} or facial expressions \cite{becattini2021plm}.

%\todo{Rifare prendendo spunto dal tomm}
%Recently, garment recommendation systems have received an increasing interest given their key role in suggesting meaningful fashion items to accommodate user personal styles and emotions. Recent examples are given by ~\cite{cheng2020fashion, song2019gp, bigi2020automatic}, where both user preferences and interest are considered to produce appropriate suggestions.

%Finally, \cite{divitiisbbb20}, propose to leverage Memory Augmented Neural Networks

\section{Overview}
\label{sec:overview}
In this paper we propose a system for compatible outfit recommendation, meaning that given a garment of a specific category (e.g., a top), we are able to propose suitable garments of a complementary category (e.g., bottoms) in order to compose an outfit.
Our system pays attention to three important aspects.
First, \textit{style-based recommendations} take into account the desired style and emotions the user would like to express. We base the recommendations on Kobayashi's Color Image Scale (CIS)~\cite{kobayashi2009color} by training a style classifier, which we integrate into our system to condition garment proposals.
Second, \textit{event-based recommendations} focus on garments that can complement the given input and at the same time can be suitable for attending a certain kind of social event, such as a work meeting, a hike or a wedding. To this end, we train an event classifier that, given an outfit, provides a probability distribution over different event categories. Similarly to the style-based recommendation, we integrate the event classifier into our system to condition the recommendations.
Third, we focus on a fundamental, yet often neglected, aspect for recommendation systems, i.e. \textit{recommendation variety}. We rely on a model specifically designed to recommend diverse garments with few redundancies and repetitions. Rather than proposing several variations of the same outfit/style, we aim at proposing different modalities that the user can choose. To this end we introduce an entropy-based evaluation to quantify such variety.

% Overall, our proposed recommendation system is shown in Fig.~\ref{fig:system}.
% \todo{immagine}
In the following we are going to provide detailed explanations of the style and event classifiers and we are then going to illustrate how these can be integrated into a recommendation system thought for recommending variegate outfits.
We carry out this study taking as reference GR-MANN~\cite{divitiisbbb20}, a recent state of the art garment recommendation system based on the usage of Memory Augmented Neural Networks.

\section{Style-based Outfit Recommendation}
\label{sec:method_style}
We refer to the task of Style-based Outfit Recommendation as the task of recommending fashion items to complement an outfit, conditioned by a given style. Styles can convey moods or emotions and with this in mind we derived an interpretation of color patterns from Kobayashi's Color Image Scale~\cite{kobayashi2009color}.
We first train a style classifier which can then be combined with a recommendation system.

\begin{figure}
    \centering
    \includegraphics[width=.9\columnwidth]{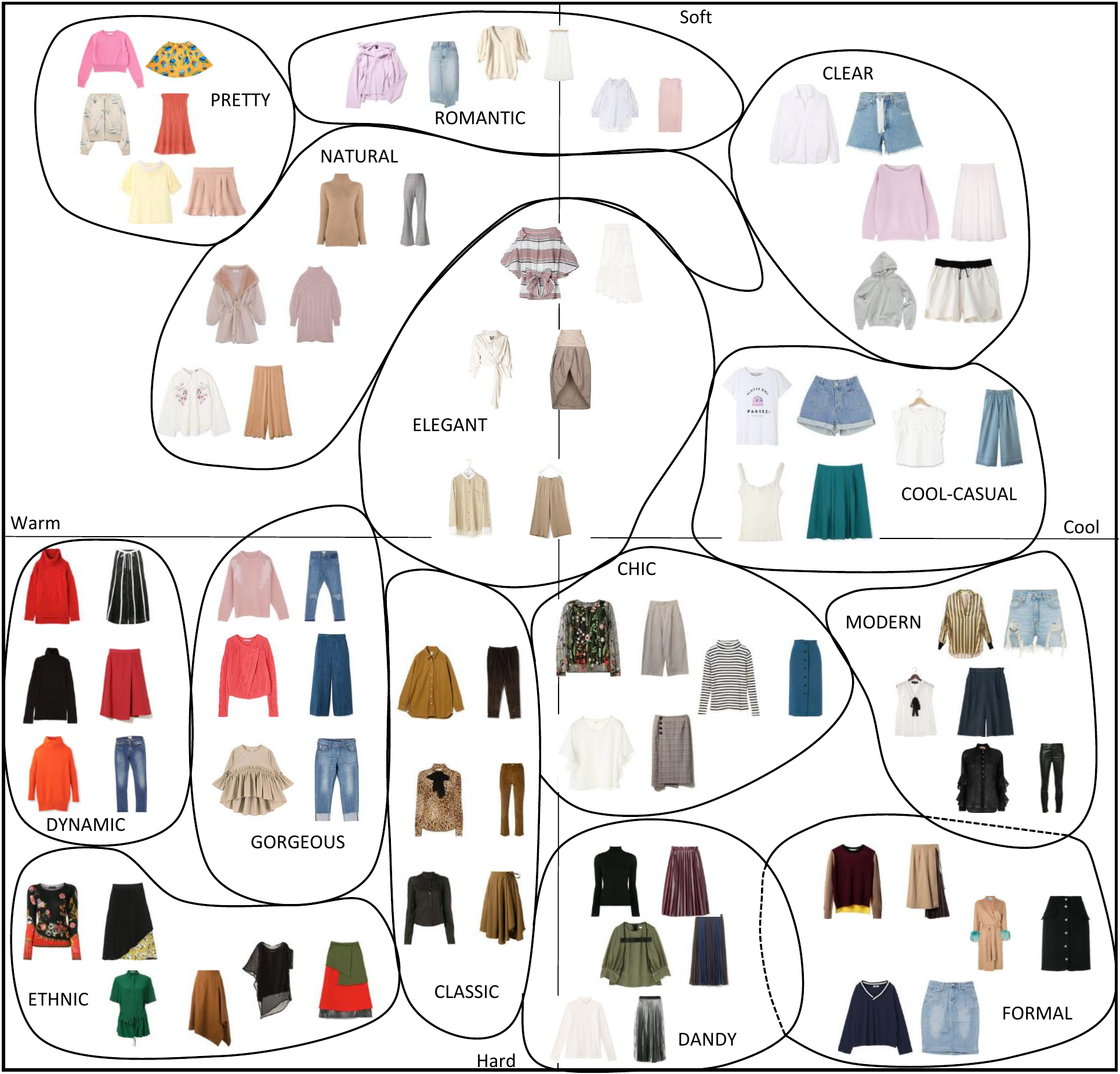}
    \caption{Kobayashi's color image scale~\cite{kobayashi2009color} applied to outfits.}
    \label{fig:kob_schema}
\end{figure}

\subsection{Kobayashi's Color Image Scale}
Originally presented in the early 90s', Kobayashi's Color Image Scale (CIS)~\cite{kobayashi2009color} connected images and colors from a psychological viewpoint, investigating associations between colors and their underlying semantics. The study carried out by Kobayashi involved a decade of color-based psychophysical experiments, asking humans to annotate triplets of colors with an adjective describing perceived emotions, moods or styles. The research eventually identified a list of 1170 triplets made of 130 unique colors and associated with 180 adjectives referred to as color images. These labels have been grouped into 15 patterns representing selected terms in fashion and lifestyle and identifying clusters in a color space spanned by two orthogonal warm-cool and short-hard axes (Fig.~\ref{fig:kob_schema}).

\begin{figure*}[t]
    \centering
    \includegraphics[trim={0 20pt 0 0},clip, width=0.45\textwidth]{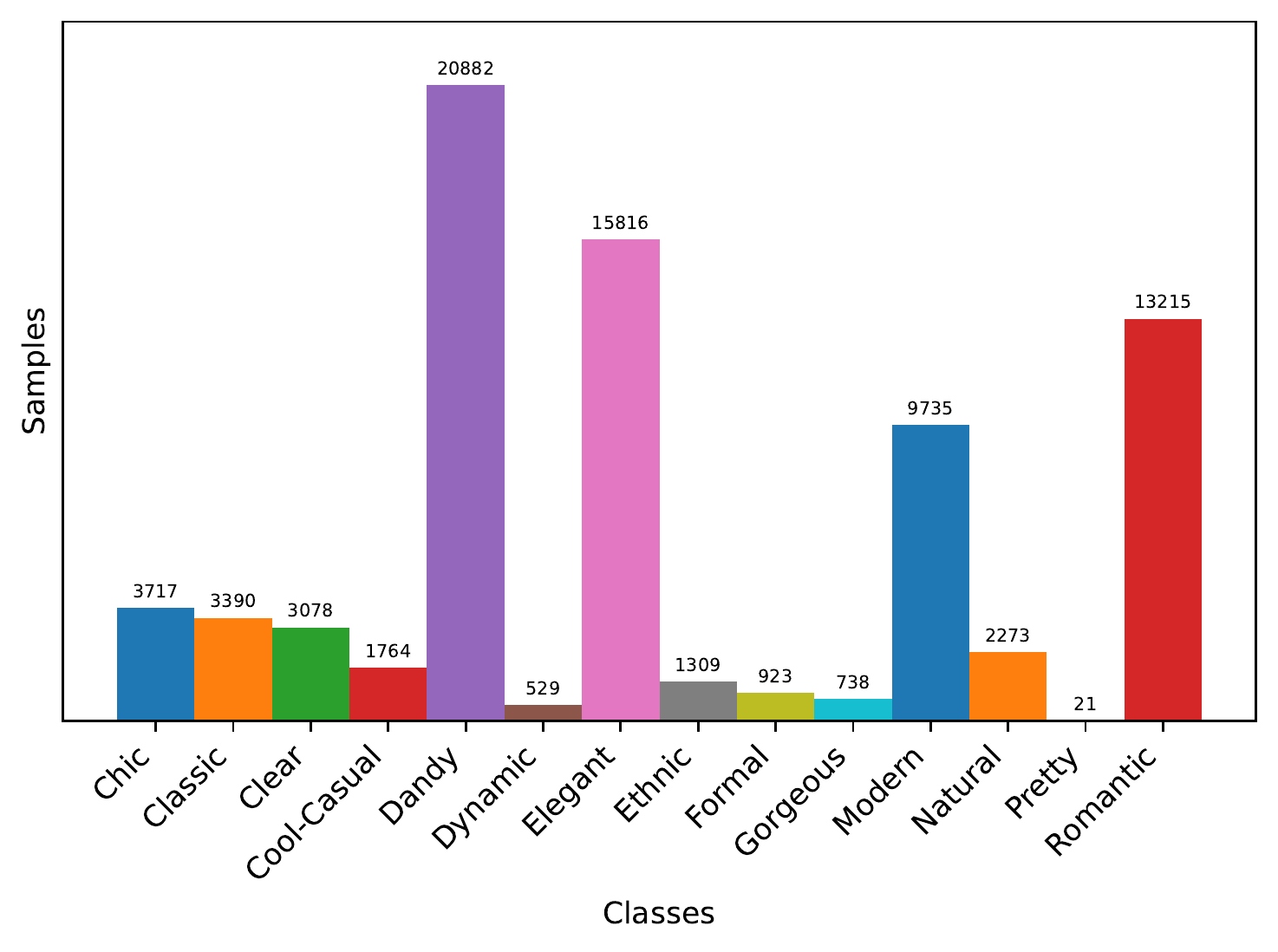}
    \includegraphics[trim={0 20pt 0 0},clip, width=0.45\textwidth]{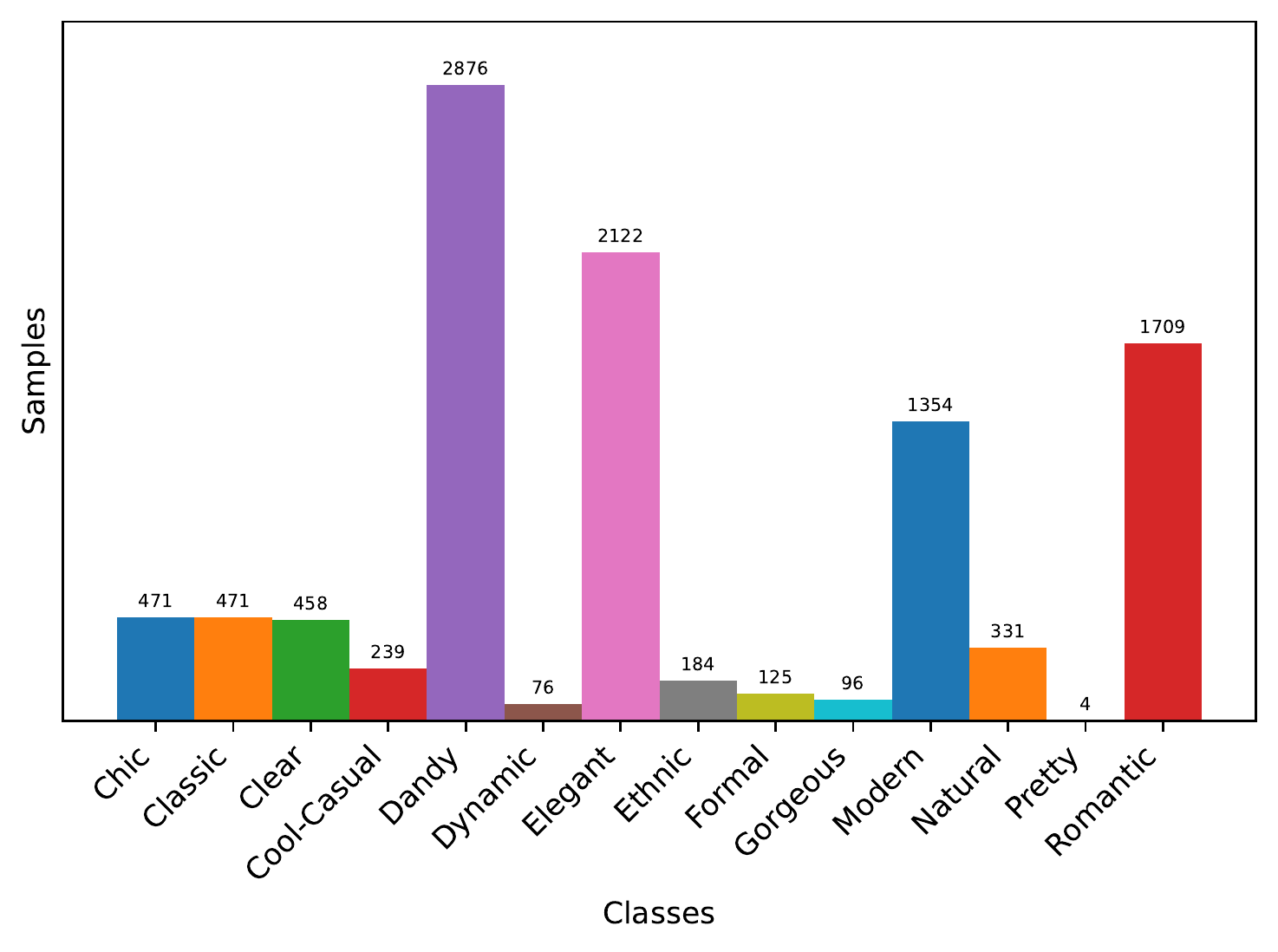}
    \caption{Sample distribution for the training (left) and test (right) sets of the collected dataset. Categories correspond to Kobayashi's color patterns.}
    \label{fig:dataset}
\end{figure*}

\subsection{Outfit Style Classifier}
We exploit a CNN model to infer the style of an outfit. In order to analyze outfits starting from a top and a bottom, we use a concatenation of the two images depicting the two garments. Styles are instead identified by Kobayashi's patterns in the Color Image Scale. We consider each pattern as a semantic description for the style of an outfit (e.g., \textit{casual}, \textit{elegant}, \textit{dandy}). Such styles indicate the feelings that an individual may want to communicate rather than describing outfit characteristics such as shape.

In order to train such a model, we collected a set of ground truth annotations following a semi-automatic procedure. Outfit images are preprocessed by removing the background and color-quantizing foreground pixels to the palette of 130 tonalities used in CIS. We then take the three most frequent colors and compare them to the Kobayashi's triplets using an euclidean distance in order to find the closest style characterizing the outfit:

\begin{equation}
d^* = \min_{p,j} \sqrt{\left \| P(c_o, p) - c_j \right \|_2}
\end{equation}

where $P(c, p)$ is the $p$-th permutation of colors in the $c$ triplet, $c_o$ is the triplet for outfit $o$ and $c_j$ the $j$-th of the 1170 triplets identified by Kobayashi.
We retain only outfits with a clear style, i.e. if $d^*<\theta$. The resulting category is then mapped to one of Kobayashi's 15 style patterns. Finally, we asked human annotators to validate the final labeling, discarding or correcting erroneous assignments.

The procedure yielded a dataset of 77390 labeled outfits for training and 10516 for testing. All outfits are taken from the IQON3000 dataset~\cite{song2019gp}.
In our experiments we considered all styles except the \textit{casual} pattern, for which enough samples are not present in the dataset, yielding to a total of 14 style categories (Fig.~\ref{fig:dataset}).
In order to recognize outfit styles, we trained a ResNet18~\cite{he2016deep} classifier that takes as input a concatenation of the top and bottom images. We trained the network for 50 epochs using an Adam optimizer with a learning rate of $0.001$.

\begin{figure*}[t]
    \centering
    \includegraphics[width=0.45\textwidth]{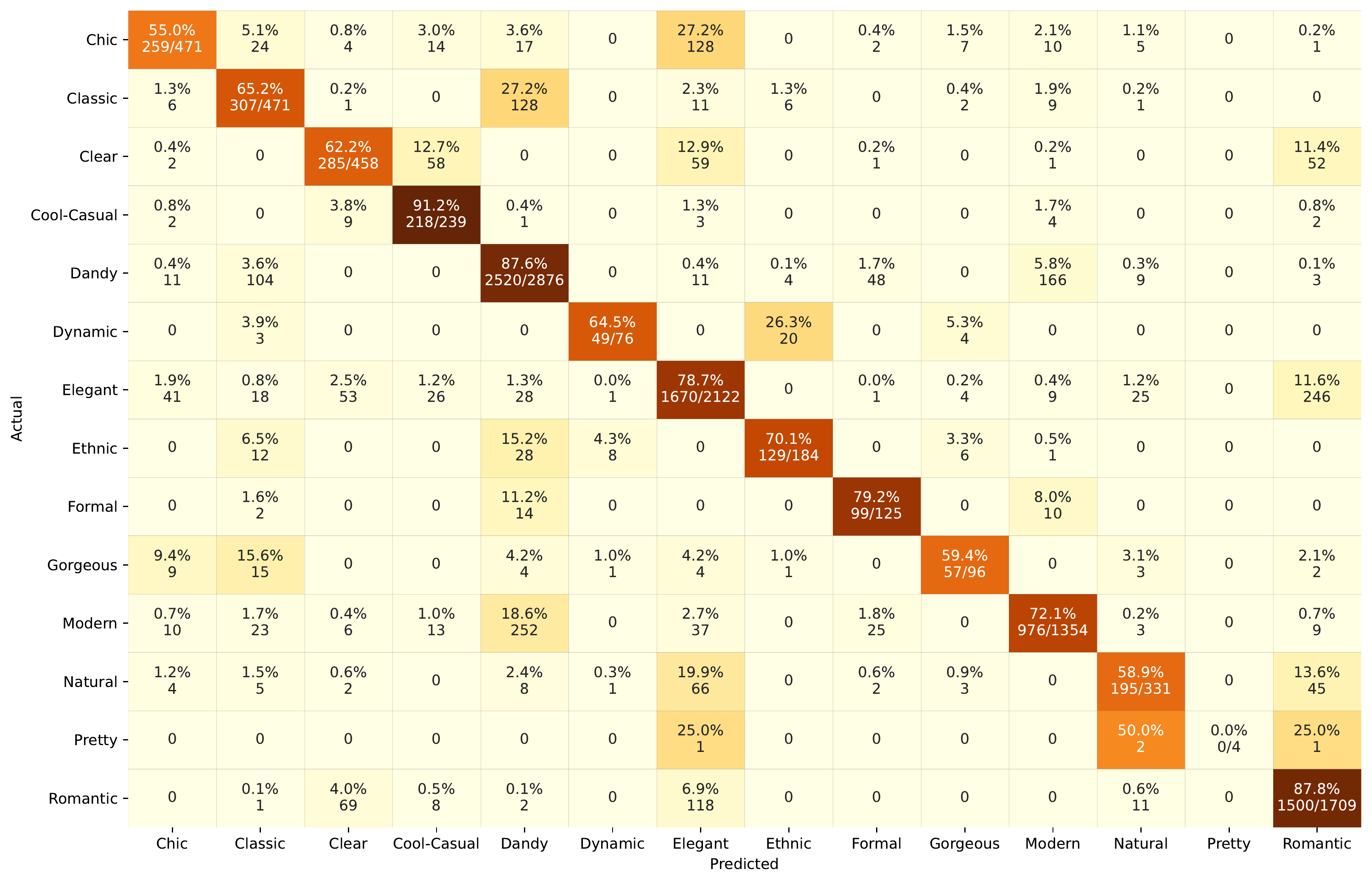}
    \includegraphics[width=0.45\textwidth]{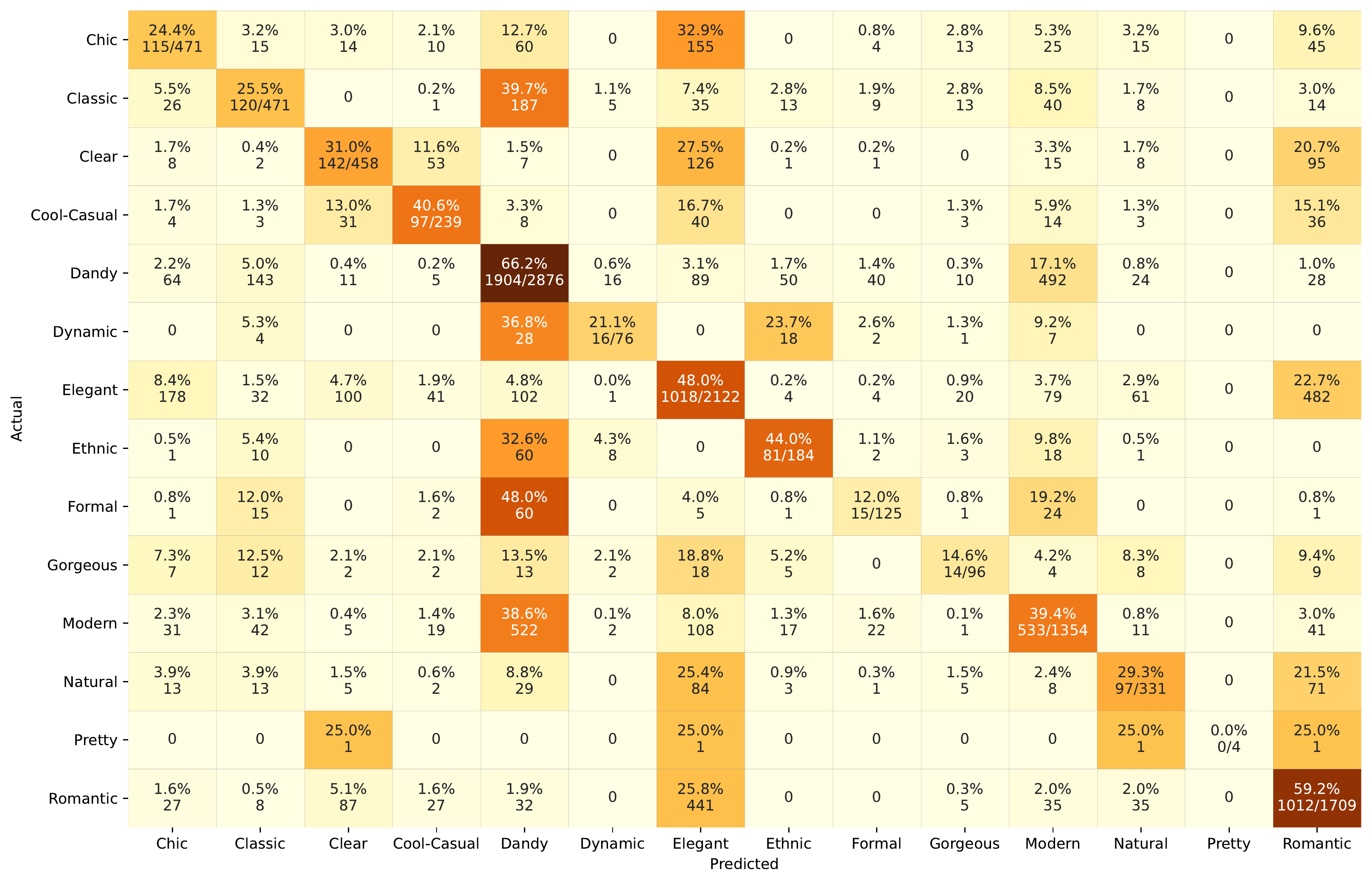}
    \caption{Confusion matrix for the outfit style classifier. Left: our ResNet18-based style classifer. Right: nearest neighbor}
    \label{fig:confmat}
\end{figure*}

\section{Event-based Outfit Recommendation}
\label{sec:method_event}

Similarly to style-based recommendation, event-based outfit recommendation aims at proposing complementary items to generate an outfit which is suitable to attend a given type of social event. Since there is no general written rule one must adhere to when it comes to dressings for social events, we first analyze images that appear in photos taken at such events. This allows us to create a dataset of garment images paired with social event labels. Such dataset, which we refer to as \textit{Fashion4Events} can then be used to train an event classifier capable of inferring a probability distribution over social event categories from a garment image.

\begin{figure*}[t]
    \centering
    \includegraphics[width=0.9\textwidth]{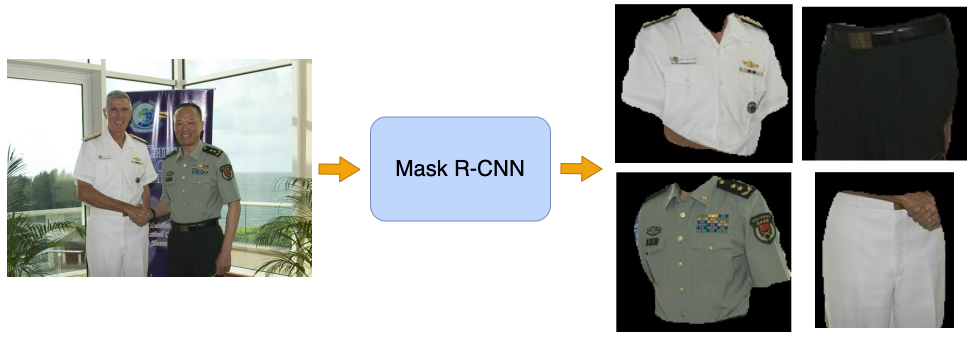}
    \caption{Garments are detected and segmented using Mask-RCNN. For each detected garment, a new image is created. Event labels associated with the original images are associated with the detected garments.}
    \label{fig:detection_garment}
\end{figure*}

\subsection{Dataset creation}
\label{sec:event_dataset}
To collect Fashion4Events, a dataset of garment images paired with social event labels, we exploited two different sources of data: the DeepFashion2 dataset \cite{ge2019deepfashion2} and the USED dataset \cite{ahmad2016used}.

DeepFashion2 is a dataset that proposed an unified benchmark for clothes detection, segmentation, retrieval and landmark prediction. It contains approximately 491K images of clothes, divided into train (391K), validation (34K) and test (67K) belonging to 13 different classes. Garments exhibit large variations in style, pose, scale, color, occlusion and viewing angle.

We use DeepFashion2 to train a garment detection and segmentation network by fine-tuning a Mask-RCNN model \cite{he2017mask}.
Once we have a model capable of detecting and segmenting individual garments in images, we apply it on the USED dataset~\cite{ahmad2016used}.
USED is a dataset comprising 525K images of people attending 14 different types of social events. Images have been  downloaded from Flickr and event types have been selected among the most common ones on social media.
%The dataset is balanced with reference to number of images per social event, i.e. the authors have collected 35K images for each category.
The categories are the following: concert, graduation, meeting, mountain-trip, picnic, sea-holiday, ski-holiday, wedding, conference, exhibition, fashion, protest, sport and theater-dance. Images represent both indoor and outdoor scenes and the dataset accounts for a variable number of people in each image.

\begin{figure*}[t]
    \centering
    \includegraphics[trim={75px 0 70px 60px},clip, width=0.49\textwidth]{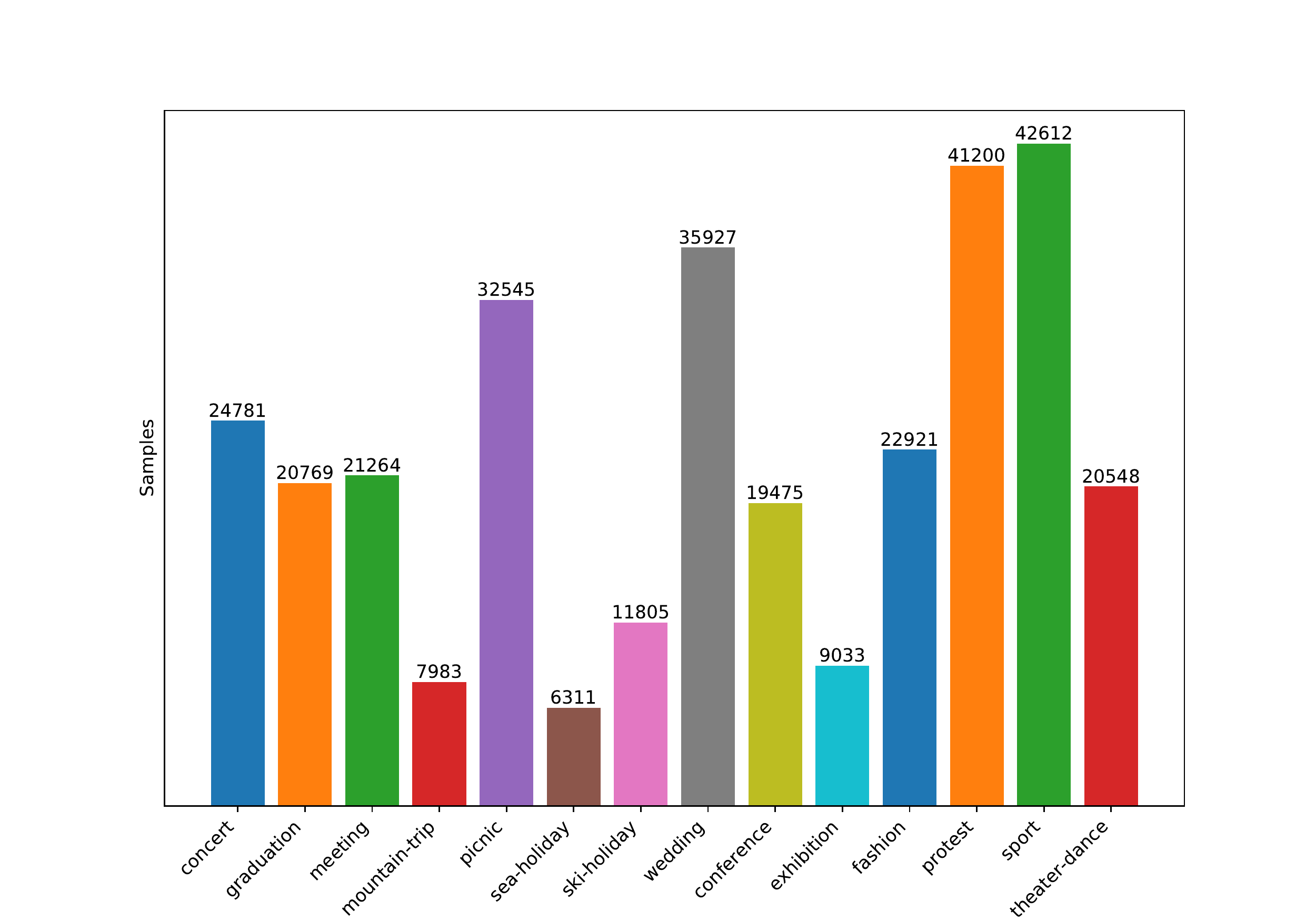}
    \includegraphics[trim={75px 0 70px 60px},clip, width=0.49\textwidth]{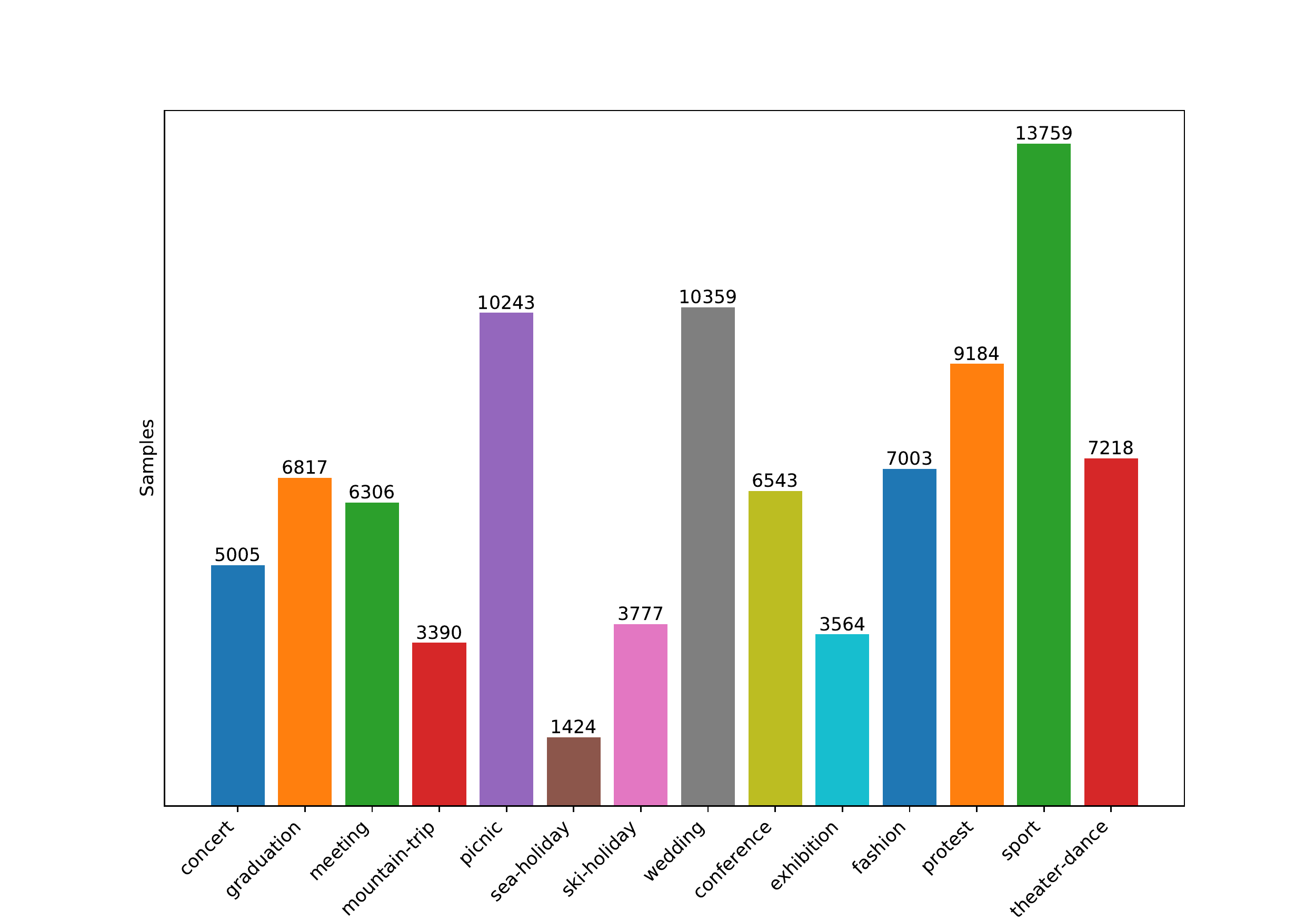}
    \caption{Sample distribution for the training (left) and test (right) sets of the collected Fashion4Events dataset. Categories correspond to social events.}
    \label{fig:dataset_event}
\end{figure*}

By applying Mask-RCNN on the images from USED, we obtain a set of detected garments paired with pixelwise segmentations and a social event label. We retain only detections with a confidence score higher than 0.8 in order to remove noise. For each detected garment, we generate a separate image by taking the segmented clothing item on a black background, as shown in Fig. \ref{fig:detection_garment}.
The final set of detections yields a dataset of 317,174 images for training and 94,592 for testing. In Fig. \ref{fig:dataset_event} we show the distribution of samples over the 14 event categories in the Fashion4Events dataset.

% \begin{figure}
%     \centering
%     \includegraphics[width=.9\columnwidth]{img/garment_event_classifier.png}
%     \caption{\todo{}}
%     \label{fig:garment_event_cls}
% \end{figure}

\subsection{Social Event Style Classification}
Thanks to the Fashion4Events dataset, which pairs segmented garment images and social event labels, we can train a garment-based event classifier. The aim of such a classifier is to provide a probability distribution of social event categories based on the style of a given garment or outfit.
We trained a ResNet18~\cite{he2016deep} model taking outfit images as inputs. As for outfit styles, we feed outfits to the model as horizontally concatenated top and bottom images. We started from a pre-trained model on ImageNet by removing the original classification head and adding two fully connected layers with respectively 512 and 14 neurons. The Adam optimizer was used to train the model, with a learning rate of 0.0005 and a weight decay of 0.0001 for 32 epochs in total.

\section{Style-Based and Event-Based Outfit Recommendation}
\label{sec:method_rec}
We integrate our outfit style classifier and outfit event classifier in the state of the art garment recommendation system GR-MANN~\cite{divitiisbbb20}. Note that in principle the classifiers could be applied to any recommendation system capable of generating an outfit composed of a top and a bottom garment.

GR-MANN is a neural network based on a persistent external memory in which non-redundant samples are stored to guide recommendations. The samples in memory express different modalities to combine tops and bottoms. At inference time, the query top is presented to the model and encoded with a convolutional encoder. Its feature is then used as key to retrieve similar tops in memory and access the correspondent stored bottoms.

We extend GR-MANN by using the outfit style and event classifiers to check whether the recommended outfits comply to a certain criterion requested by the user. Once a bottom is proposed, it is concatenated to the input top and fed to the classifiers. We perform different fusion methods, in order to characterize the behaviour of the system, as detailed in Sec.~\ref{sec:exp}.

The choice of exploiting GR-MANN stems from the fact that its memory is populated by a controller, trained to store only relevant samples. Classifying such samples based on style and event type will therefore allow us to quantify the diversity that GR-MANN strives to achieve in its recommendations.

%\todo{controllare nome dei classificatori}

\section{Experiments}
\label{sec:exp}
We demonstrate our method on the IQON3000 dataset~\cite{song2019gp}, performing several evaluations. At first we discuss the accuracy of the style classifier and the event classifier and then we combine them with GR-MANN to analyze their capability to recommend a diverse set of bottom garments covering multiple styles. In order to evaluate the performance of the event classifier we rely on the event-garment dataset based on USED, obtained as explained in Sec.~\ref{sec:event_dataset}.

\subsection{Style Classifier Evaluation}

In Fig.~\ref{fig:confmat} the confusion matrix for the style classifier is shown. As a reference, a Nearest Neighbor (NN) baseline is also reported. Here we simply extract features from a standard ResNet, pretrained on ImageNet, and transfer the style category from the closest sample in the training set onto a given test outfit. Our model achieves an accuracy of 78.56\% while the NN baseline only 49.11\%, highlighting how the task is not as straightforward as comparing visual features.

Interestingly, most of the errors committed by the model tend to confuse similar categories, such as \textit{elegant} and \textit{chic} or \textit{classic} and \textit{dandy}.

\begin{table}[]
\caption{Accuracy and mAP obtained by GR-MANN~\cite{divitiisbbb20} on the IQON3000~\cite{song2019gp} dataset. The metrics are computed in order to retrieve an outfit with the same style of the ground truth.}
\label{tab:acc_map_exp1}
\centering
\resizebox{\columnwidth}{!}{%
\let\b\textbf
\begin{tabular}{l|ccccccc}
Num Items   & 5     & 10    & 20    & 30    & 40    & 50    & 60    
\\ \hline
Accuracy    & 73.42 & 84.17 & 91.47 & 94.22 & 95.73 & 96.70 & 97.39 \\
Random Acc. & 36.50 & 53.12 & 80.37 & 91.14 & 96.51 & 99.02 & 99.89 \\
\hline
mAP         & 48.17 & 46.35 & 43.46 & 41.78 & 40.70 & 39.96 & 39.40 \\
Random mAP  & 18.14 & 17.21 & 17.04 & 16.59 & 14.39 & 13.41 & 11.46  
\end{tabular}%
}
\end{table}

\subsection{Event Classifier Evaluation}

\begin{figure*}[t]
    \centering
    \includegraphics[width=0.49\textwidth]{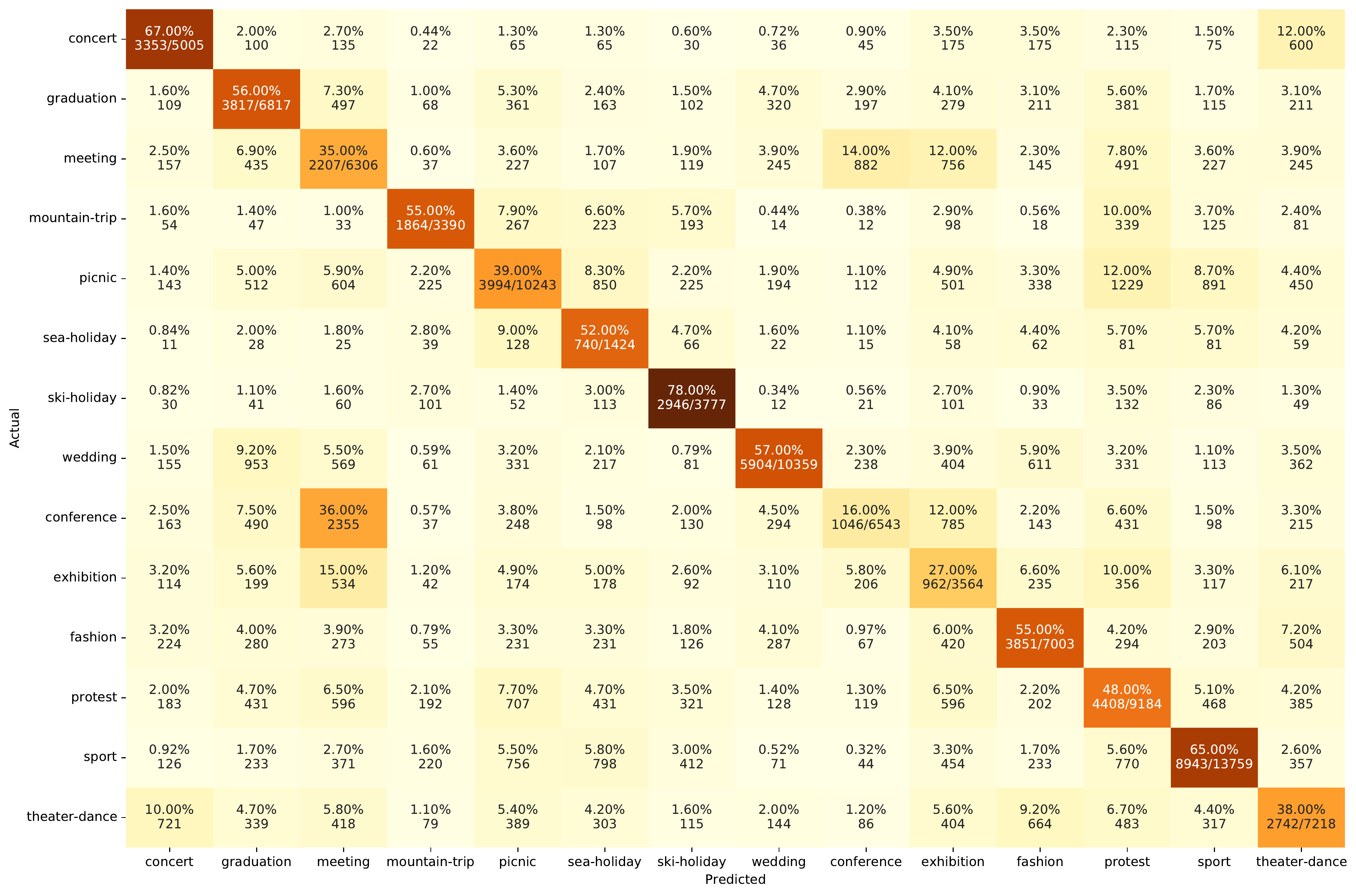}
    \includegraphics[width=0.49\textwidth]{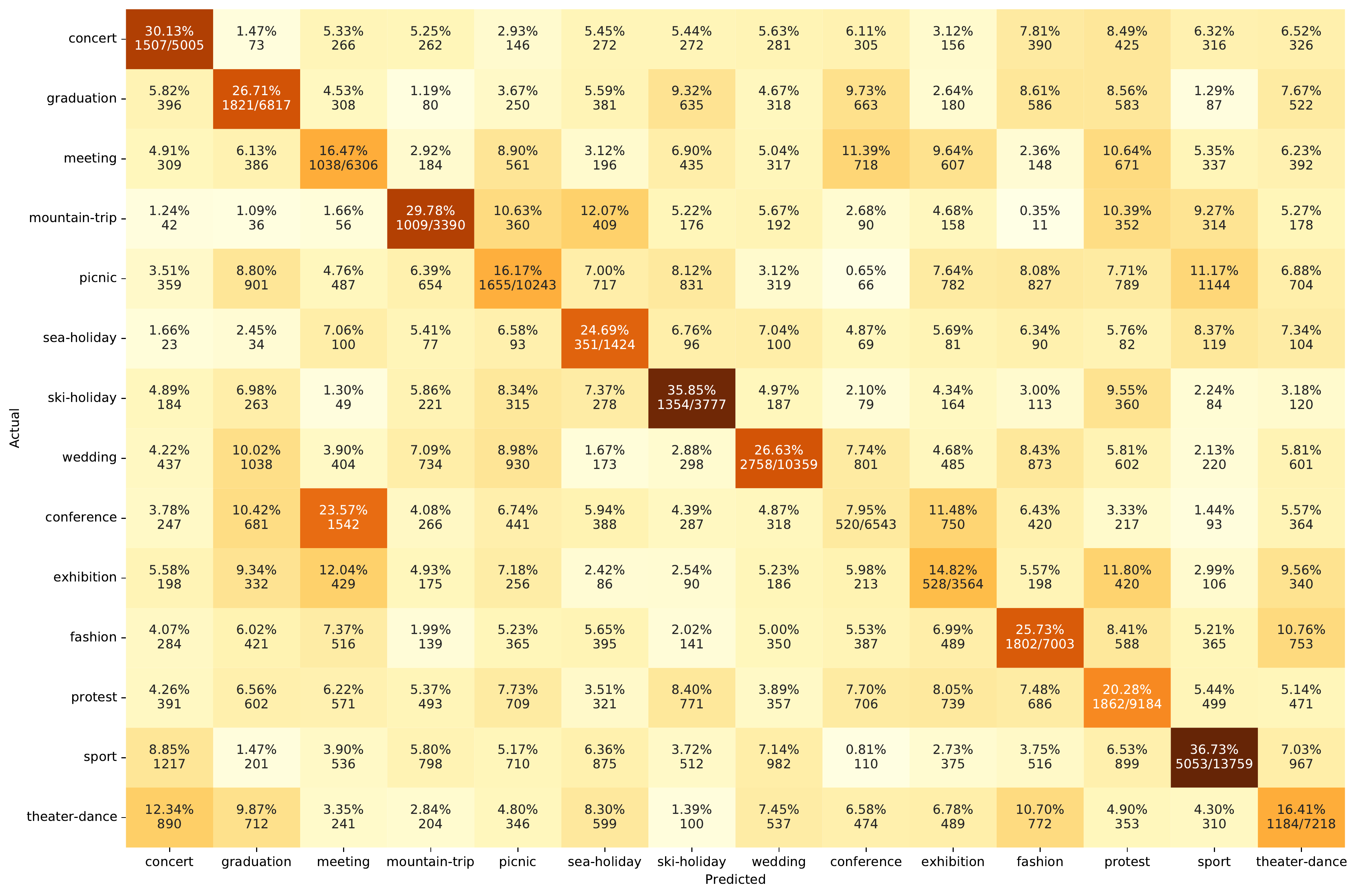}
    \caption{Confusion matrix for the outfit style classifier. Left: our ResNet18-based style classifier. Right: Nearest Neighbor.}
    \label{fig:confmatevent}
\end{figure*}

Overall, the event classifier obtained a 49.9\% accuracy on the test set of the garment-event dataset (Sec.~\ref{sec:event_dataset}).
As can be seen in the confusion matrix in Fig.~\ref{fig:confmatevent}, the most difficult event class appears to be \textit{Conference}, which is often confused with \textit{Meeting}. This is easily explainable given that these two social events are work related and thus have similar dress-codes.  In the Nearest Neighbor baseline, the accuracy drops to 23.45\% and the confusion among related classes is more present, such as between \textit{Theatre-dance} and \textit{Concert} or \textit{Exhibition} and \textit{Meeting}.

\subsection{Style-Based and Event-Based Outfit Recommendation Evaluation}

We now assess the capabilities of the GR-MANN recommendation system with reference to styles and social events.
First, we measure how the recommendation system is able to suggest garments that comply with the ground truth style and event category, without adding any prior knowledge to the model.
We report Accuracy, measuring if at least one of the recommended outfits adheres with the requested category, and mAP, which takes the ranking of the correct bottoms into account.

In Tab.~\ref{tab:acc_map_exp1} we compare the results for style-based recommendation obtained by the model against a baseline in which style categories are drawn at random. It can be seen that, both for Accuracy and for mAP, the results improve considerably and that the model is able to provide at least an outfit with the desired style most of the times even with only 5 recommendations.

As for event-based recommendation, we report a similar analysis in Tab.~\ref{tab:acc_map_exp1_event}. Also in this case, the model generates a set of recommendations comprising a suitable bottom to comply with the social event of interest.

\begin{table}[]
\caption{Accuracy and mAP obtained by GR-MANN~\cite{divitiisbbb20} on the IQON3000~\cite{song2019gp} dataset. The metrics are computed in order to retrieve an outfit with the same social event category of the ground truth.}
\label{tab:acc_map_exp1_event}
\centering
\resizebox{\columnwidth}{!}{%
\let\b\textbf
\begin{tabular}{l|ccccccc}
Num Items   & 5     & 10    & 20    & 30    & 40    & 50    & 60    
\\ \hline
Accuracy    & 75.83 & 88.50 & 94.96 & 95.06 & 96.72 & 97.24 & 97.70 \\
Random Acc. & 37.12 & 53.03 & 79.87 & 91.21 & 96.73 & 98.99 & 99.16 \\
\hline
mAP         & 43.05 & 40.00 & 36.66 & 34.25 & 33.07 & 32.30 & 31.42 \\
Random mAP  & 18.31 & 16.87 & 17.34 & 16.44 & 14.12 & 13.36 & 11.07  
\end{tabular}%
}
\end{table}

Additionally, following the evaluation protocol of~\cite{divitiisbbb20}, we also measure accuracy color-wise, category-wise and combining both together. However, we filter the output of GR-MANN in order to provide a ranked list of bottoms with the correct category using the style classifier or the event classifier. Therefore, in this experiment we are relaxing the formulation of the task, assuming that the desired style is known a-priori.
The rationale behind this evaluation is to see if the proposed garments share similar visual traits with the ground truth when performing a category-conditioned (either style-based or event-based) recommendation.
In Tab.~\ref{tab:acc_map_exp2} and Tab.~\ref{tab:acc_map_exp2_event} we report results for both accuracy and mAP.

\begin{table}[]
\caption{Accuracy and mAP varying the number or retrieved items. All proposals are filtered by the style classifier in order to share the desired one.}
\label{tab:acc_map_exp2}
\centering
\resizebox{\columnwidth}{!}{%
\let\b\textbf
\begin{tabular}{l|ccccccc}
Num Items             & 5     & 10    & 20    & 30    & 40    & 50    & 60    \\
\hline
Cat $\times$ Col Acc. & 57.66 & 59.22 & 61.92 & 64.37 & 66.42 & 68.36 & 69.97 \\
Category Accuracy     & 83.92 & 84.81 & 86.27 & 87.60 & 88.68 & 89.67 & 90.39 \\
Color Accuracy        & 81.41 & 82.73 & 84.71 & 86.37 & 87.69 & 88.82 & 89.77 \\
\hline
mAP                   & 18.50 & 18.48 & 18.37 & 18.23 & 18.11 & 17.98 & 17.88 \\
%Random mAP            & 18.42 & 18.20 & 18.12 & 17.98 & 17.79 & 17.78 & 17.82
\end{tabular}%
}
\end{table}

\begin{table}[]
\caption{Accuracy and mAP varying the number or retrieved items. All proposals are filtered by the event classifier in order to share the desired category.}
\label{tab:acc_map_exp2_event}
\centering
\resizebox{\columnwidth}{!}{%
\let\b\textbf
\begin{tabular}{l|ccccccc}
Num Items             & 5     & 10    & 20    & 30    & 40    & 50    & 60    \\
\hline
Cat $\times$ Col Acc. & 52.00 & 53.70 & 55.46 & 57.92 & 60.86 & 62.74 & 64.68 \\
Category Accuracy     & 86.14 & 86.36 & 88.24 & 88.78 & 89.98 & 90.76 & 90.40 \\
Color Accuracy        & 75.92 & 77.70 & 79.96 & 81.58 & 83.78 & 84.74 & 86.46 \\
\hline
mAP                   & 16.02 & 15.92 & 15.47 & 15.33 & 15.57 & 15.42 & 14.98 \\
%Random mAP            & 18.42 & 18.20 & 18.12 & 17.98 & 17.79 & 17.78 & 17.82
\end{tabular}%
}
\end{table}

\subsection{Recommendation Diversity Evaluation}
As studied in \cite{wang2006}, diversity is an important aspect of information retrieval systems. To this end, we also perform an evaluation of the entropy of the proposed labels to establish the variation degree of our proposals. This evaluation was first proposed in \cite{cuffaro2016segmentation} to perform an unsupervised assessment of a generic classifier with unsupervised data. Given a probability distribution $X = \{x_1, x_2, \dots, x_n\}$ over $N$ different classes, we can compute the Shannon entropy $H$ for the probability vector $X$ as $H(X) = -\sum_{i=1}^{N}x_{i}log(x_i)$. The entropy will be $0$ when all samples are labeled with the same class, and will increase as more information and diversity are introduced in the predictions. Ideally we would like to stay as close as possible to the entropy of any random label permutations, but preserving good recommendation results. Results, shown in Tab.~\ref{tab:entropy} and Tab.~\ref{tab:entropy_event}, show that our method is able to maintain a reasonable amount of entropy in the predictions while performing significantly better than random, as shown in Tab.~\ref{tab:acc_map_exp1} and Tab.~\ref{tab:acc_map_exp1_event}.

Interestingly, the entropy for Kobayashi's style categories is higher than the one relative to event categories. We attribute this to the distribution of garments in IQON3000, for which certain categories such as ski-holiday and wedding are underrepresented. 

\begin{table}[]
\caption{Entropy of the recommendations with reference to outfit styles. A sufficiently high entropy indicates variety in the proposed outfits.}
\label{tab:entropy}
\centering
\resizebox{\columnwidth}{!}{%
\let\b\textbf
\newcommand{\tbs}{\textbackslash}
\begin{tabular}{l|ccccccc}
Method \tbs Num Items & 5     & 10    & 20    & 30    & 40    & 50    & 60    \\
\hline
% Uniform               & 1.609 & 2.303 & 2.580 & 2.627 & 2.631 & 2.629 & 2.634 \\
Random                & 1.419 & 1.905 & 2.255 & 2.391 & 2.458 & 2.499 & 2.523 \\
Style-Entropy         & 0.849 & 1.061 & 1.152 & 1.265 & 1.267 & 1.288 & 1.317
\end{tabular}%
}
\end{table}

\begin{table}[]
\caption{Entropy of the recommendations with reference to outfit event categories. A sufficiently high entropy indicates variety in the proposed outfits.}
\label{tab:entropy_event}
\centering
\resizebox{\columnwidth}{!}{%
\let\b\textbf
\newcommand{\tbs}{\textbackslash}
\begin{tabular}{l|ccccccc}
Method \tbs Num Items & 5     & 10    & 20    & 30    & 40    & 50    & 60    \\
\hline
% Uniform               & 1.609 & 2.303 & 2.580 & 2.627 & 2.631 & 2.629 & 2.634 \\
Random                & 1.353 & 1.839 & 2.189 & 2.325 & 2.392 & 2.433 & 2.457 \\
Event-Entropy         & 0.671 & 0.819 & 0.908 & 0.941 & 0.958 & 0.965 & 0.970
\end{tabular}%
}
\end{table}

\section{Conclusions}
\label{sec:conclusions}
In this paper we presented an approach to take into account a style based filtering and an event based filtering for fashion recommendation. Styles can convey a color-based mood according to Kobayashi's Color Image Scale or can reflect dress codes for specicic social event categories. We leveraged the work of Kobayashi to train a style classifier that we used to filter the results of a memory network based garment recommender. Similarly, we trained a garment-based event classifier to be combined with the recommender by exploiting a garment detector and Fashion4Events, an image dataset of annotated social events.
Experiments show that our system is able to generalise on color styles and social events and that the recommendation system is able to propose a variety of outfit styles compatible with the query garment.

\section*{Data Availability}
The datasets generated and analysed during the current study are available in the Fashion4Events repository: \url{https://github.com/fedebecat/Fashion4Events}.

\section*{Conflict of Interest}
The authors declare that they have no conflict of interest.

\section*{Acknowledgments}
This work was partially supported by the Italian MIUR within PRIN 2017, Project Grant 20172BH297: I-MALL - improving the customer experience in stores by intelligent computer vision.


\bibliography{sn-bibliography}% common bib file


\begin{thebibliography}{32}
\ifx \bisbn   \undefined \def \bisbn  #1{ISBN #1}\fi
\ifx \binits  \undefined \def \binits#1{#1}\fi
\ifx \bauthor  \undefined \def \bauthor#1{#1}\fi
\ifx \batitle  \undefined \def \batitle#1{#1}\fi
\ifx \bjtitle  \undefined \def \bjtitle#1{#1}\fi
\ifx \bvolume  \undefined \def \bvolume#1{\textbf{#1}}\fi
\ifx \byear  \undefined \def \byear#1{#1}\fi
\ifx \bissue  \undefined \def \bissue#1{#1}\fi
\ifx \bfpage  \undefined \def \bfpage#1{#1}\fi
\ifx \blpage  \undefined \def \blpage #1{#1}\fi
\ifx \burl  \undefined \def \burl#1{\textsf{#1}}\fi
\ifx \doiurl  \undefined \def \doiurl#1{\url{https://doi.org/#1}}\fi
\ifx \betal  \undefined \def \betal{\textit{et al.}}\fi
\ifx \binstitute  \undefined \def \binstitute#1{#1}\fi
\ifx \binstitutionaled  \undefined \def \binstitutionaled#1{#1}\fi
\ifx \bctitle  \undefined \def \bctitle#1{#1}\fi
\ifx \beditor  \undefined \def \beditor#1{#1}\fi
\ifx \bpublisher  \undefined \def \bpublisher#1{#1}\fi
\ifx \bbtitle  \undefined \def \bbtitle#1{#1}\fi
\ifx \bedition  \undefined \def \bedition#1{#1}\fi
\ifx \bseriesno  \undefined \def \bseriesno#1{#1}\fi
\ifx \blocation  \undefined \def \blocation#1{#1}\fi
\ifx \bsertitle  \undefined \def \bsertitle#1{#1}\fi
\ifx \bsnm \undefined \def \bsnm#1{#1}\fi
\ifx \bsuffix \undefined \def \bsuffix#1{#1}\fi
\ifx \bparticle \undefined \def \bparticle#1{#1}\fi
\ifx \barticle \undefined \def \barticle#1{#1}\fi
\bibcommenthead
\ifx \bconfdate \undefined \def \bconfdate #1{#1}\fi
\ifx \botherref \undefined \def \botherref #1{#1}\fi
\ifx \url \undefined \def \url#1{\textsf{#1}}\fi
\ifx \bchapter \undefined \def \bchapter#1{#1}\fi
\ifx \bbook \undefined \def \bbook#1{#1}\fi
\ifx \bcomment \undefined \def \bcomment#1{#1}\fi
\ifx \oauthor \undefined \def \oauthor#1{#1}\fi
\ifx \citeauthoryear \undefined \def \citeauthoryear#1{#1}\fi
\ifx \endbibitem  \undefined \def \endbibitem {}\fi
\ifx \bconflocation  \undefined \def \bconflocation#1{#1}\fi
\ifx \arxivurl  \undefined \def \arxivurl#1{\textsf{#1}}\fi
\csname PreBibitemsHook\endcsname

%%% 1
\bibitem{kobayashi2009color}
\begin{botherref}
\oauthor{\bsnm{Kobayashi}, \binits{S.}}:
Color image scale.
http://www. ncd-ri. co. jp/english/main\_0104. html
(2009)
\end{botherref}
\endbibitem

%%% 2
\bibitem{wang2006}
\begin{bchapter}
\bauthor{\bsnm{Wang}, \binits{J.Z.}},
\bauthor{\bsnm{Boujemaa}, \binits{N.}},
\bauthor{\bsnm{Del~Bimbo}, \binits{A.}},
\bauthor{\bsnm{Geman}, \binits{D.}},
\bauthor{\bsnm{Hauptmann}, \binits{A.G.}},
\bauthor{\bsnm{Tesi\'{c}}, \binits{J.}}:
\bctitle{Diversity in multimedia information retrieval research}.
In: \bbtitle{Proceedings of the 8th ACM International Workshop on Multimedia
  Information Retrieval}.
\bsertitle{MIR '06},
pp. \bfpage{5}--\blpage{12}.
\bpublisher{Association for Computing Machinery},
\blocation{New York, NY, USA}
(\byear{2006}).
\doiurl{10.1145/1178677.1178681}.
\burl{https://doi.org/10.1145/1178677.1178681}
\end{bchapter}
\endbibitem

%%% 3
\bibitem{de2021style}
\begin{bchapter}
\bauthor{\bsnm{De~Divitiis}, \binits{L.}},
\bauthor{\bsnm{Becattini}, \binits{F.}},
\bauthor{\bsnm{Baecchi}, \binits{C.}},
\bauthor{\bsnm{Del~Bimbo}, \binits{A.}}:
\bctitle{Style-based outfit recommendation}.
In: \bbtitle{2021 International Conference on Content-Based Multimedia Indexing
  (CBMI)},
pp. \bfpage{1}--\blpage{4}
(\byear{2021}).
\bcomment{IEEE}
\end{bchapter}
\endbibitem

%%% 4
\bibitem{song2019gp}
\begin{bchapter}
\bauthor{\bsnm{Song}, \binits{X.}},
\bauthor{\bsnm{Han}, \binits{X.}},
\bauthor{\bsnm{Li}, \binits{Y.}},
\bauthor{\bsnm{Chen}, \binits{J.}},
\bauthor{\bsnm{Xu}, \binits{X.-S.}},
\bauthor{\bsnm{Nie}, \binits{L.}}:
\bctitle{Gp-bpr: Personalized compatibility modeling for clothing matching}.
In: \bbtitle{Proceedings of the 27th ACM International Conference on
  Multimedia},
pp. \bfpage{320}--\blpage{328}
(\byear{2019})
\end{bchapter}
\endbibitem

%%% 5
\bibitem{sarkar2022outfittransformer}
\begin{bchapter}
\bauthor{\bsnm{Sarkar}, \binits{R.}},
\bauthor{\bsnm{Bodla}, \binits{N.}},
\bauthor{\bsnm{Vasileva}, \binits{M.}},
\bauthor{\bsnm{Lin}, \binits{Y.-L.}},
\bauthor{\bsnm{Beniwal}, \binits{A.}},
\bauthor{\bsnm{Lu}, \binits{A.}},
\bauthor{\bsnm{Medioni}, \binits{G.}}:
\bctitle{Outfittransformer: Outfit representations for fashion recommendation}.
In: \bbtitle{Proceedings of the IEEE/CVF Conference on Computer Vision and
  Pattern Recognition},
pp. \bfpage{2263}--\blpage{2267}
(\byear{2022})
\end{bchapter}
\endbibitem

%%% 6
\bibitem{de2022disentangling}
\begin{botherref}
\oauthor{\bsnm{De~Divitiis}, \binits{L.}},
\oauthor{\bsnm{Becattini}, \binits{F.}},
\oauthor{\bsnm{Baecchi}, \binits{C.}},
\oauthor{\bsnm{Bimbo}, \binits{A.D.}}:
Disentangling features for fashion recommendation.
ACM Transactions on Multimedia Computing, Communications, and Applications
  (TOMM)
(2022)
\end{botherref}
\endbibitem

%%% 7
\bibitem{baldrati2021conditioned}
\begin{bchapter}
\bauthor{\bsnm{Baldrati}, \binits{A.}},
\bauthor{\bsnm{Bertini}, \binits{M.}},
\bauthor{\bsnm{Uricchio}, \binits{T.}},
\bauthor{\bsnm{Del~Bimbo}, \binits{A.}}:
\bctitle{Conditioned image retrieval for fashion using contrastive learning and
  clip-based features}.
In: \bbtitle{ACM Multimedia Asia},
pp. \bfpage{1}--\blpage{5}
(\byear{2021})
\end{bchapter}
\endbibitem

%%% 8
\bibitem{divitiisbbb20}
\begin{bchapter}
\bauthor{\bsnm{Divitiis}, \binits{L.D.}},
\bauthor{\bsnm{Becattini}, \binits{F.}},
\bauthor{\bsnm{Baecchi}, \binits{C.}},
\bauthor{\bsnm{Bimbo}, \binits{A.D.}}:
\bctitle{Garment recommendation with memory augmented neural networks}.
In: \bbtitle{Pattern Recognition. {ICPR} International Workshops and Challenges
  - Virtual Event, January 10-15, 2021, Proceedings, Part {II}}.
\bsertitle{Lecture Notes in Computer Science},
vol. \bseriesno{12662},
pp. \bfpage{282}--\blpage{295}.
\bpublisher{Springer}, \blocation{???}
(\byear{2020}).
\doiurl{10.1007/978-3-030-68790-8\_23}.
\burl{https://doi.org/10.1007/978-3-030-68790-8\_23}
\end{bchapter}
\endbibitem

%%% 9
\bibitem{sa2022diversity}
\begin{bchapter}
\bauthor{\bsnm{S{\'a}}, \binits{J.}},
\bauthor{\bsnm{Queiroz~Marinho}, \binits{V.}},
\bauthor{\bsnm{Magalh{\~a}es}, \binits{A.R.}},
\bauthor{\bsnm{Lacerda}, \binits{T.}},
\bauthor{\bsnm{Goncalves}, \binits{D.}}:
\bctitle{Diversity vs relevance: A practical multi-objective study in luxury
  fashion recommendations}.
In: \bbtitle{Proceedings of the 45th International ACM SIGIR Conference on
  Research and Development in Information Retrieval},
pp. \bfpage{2405}--\blpage{2409}
(\byear{2022})
\end{bchapter}
\endbibitem

%%% 10
\bibitem{hou2021learning}
\begin{bchapter}
\bauthor{\bsnm{Hou}, \binits{Y.}},
\bauthor{\bsnm{Vig}, \binits{E.}},
\bauthor{\bsnm{Donoser}, \binits{M.}},
\bauthor{\bsnm{Bazzani}, \binits{L.}}:
\bctitle{Learning attribute-driven disentangled representations for interactive
  fashion retrieval}.
In: \bbtitle{Proceedings of the IEEE/CVF International Conference on Computer
  Vision},
pp. \bfpage{12147}--\blpage{12157}
(\byear{2021})
\end{bchapter}
\endbibitem

%%% 11
\bibitem{vasileva2018learning}
\begin{bchapter}
\bauthor{\bsnm{Vasileva}, \binits{M.I.}},
\bauthor{\bsnm{Plummer}, \binits{B.A.}},
\bauthor{\bsnm{Dusad}, \binits{K.}},
\bauthor{\bsnm{Rajpal}, \binits{S.}},
\bauthor{\bsnm{Kumar}, \binits{R.}},
\bauthor{\bsnm{Forsyth}, \binits{D.}}:
\bctitle{Learning type-aware embeddings for fashion compatibility}.
In: \bbtitle{Proceedings of the European Conference on Computer Vision (ECCV)},
pp. \bfpage{390}--\blpage{405}
(\byear{2018})
\end{bchapter}
\endbibitem

%%% 12
\bibitem{song2018neural}
\begin{bchapter}
\bauthor{\bsnm{Song}, \binits{X.}},
\bauthor{\bsnm{Feng}, \binits{F.}},
\bauthor{\bsnm{Han}, \binits{X.}},
\bauthor{\bsnm{Yang}, \binits{X.}},
\bauthor{\bsnm{Liu}, \binits{W.}},
\bauthor{\bsnm{Nie}, \binits{L.}}:
\bctitle{Neural compatibility modeling with attentive knowledge distillation}.
In: \bbtitle{The 41st International ACM SIGIR Conference on Research \&
  Development in Information Retrieval},
pp. \bfpage{5}--\blpage{14}
(\byear{2018})
\end{bchapter}
\endbibitem

%%% 13
\bibitem{liu2019neural}
\begin{barticle}
\bauthor{\bsnm{Liu}, \binits{J.}},
\bauthor{\bsnm{Song}, \binits{X.}},
\bauthor{\bsnm{Chen}, \binits{Z.}},
\bauthor{\bsnm{Ma}, \binits{J.}}:
\batitle{Neural fashion experts: I know how to make the complementary clothing
  matching}.
\bjtitle{Neurocomputing}
\bvolume{359},
\bfpage{249}--\blpage{263}
(\byear{2019})
\end{barticle}
\endbibitem

%%% 14
\bibitem{sagar2020pai}
\begin{bchapter}
\bauthor{\bsnm{Sagar}, \binits{D.}},
\bauthor{\bsnm{Garg}, \binits{J.}},
\bauthor{\bsnm{Kansal}, \binits{P.}},
\bauthor{\bsnm{Bhalla}, \binits{S.}},
\bauthor{\bsnm{Shah}, \binits{R.R.}},
\bauthor{\bsnm{Yu}, \binits{Y.}}:
\bctitle{Pai-bpr: Personalized outfit recommendation scheme with attribute-wise
  interpretability}.
In: \bbtitle{2020 IEEE Sixth International Conference on Multimedia Big Data
  (BigMM)},
pp. \bfpage{221}--\blpage{230}
(\byear{2020}).
\bcomment{IEEE}
\end{bchapter}
\endbibitem

%%% 15
\bibitem{graves2014neural}
\begin{botherref}
\oauthor{\bsnm{Graves}, \binits{A.}},
\oauthor{\bsnm{Wayne}, \binits{G.}},
\oauthor{\bsnm{Danihelka}, \binits{I.}}:
Neural turing machines.
arXiv preprint arXiv:1410.5401
(2014)
\end{botherref}
\endbibitem

%%% 16
\bibitem{weston2014memory}
\begin{botherref}
\oauthor{\bsnm{Weston}, \binits{J.}},
\oauthor{\bsnm{Chopra}, \binits{S.}},
\oauthor{\bsnm{Bordes}, \binits{A.}}:
Memory networks.
arXiv preprint arXiv:1410.3916
(2014)
\end{botherref}
\endbibitem

%%% 17
\bibitem{marchetti2020memnet}
\begin{bchapter}
\bauthor{\bsnm{Marchetti}, \binits{F.}},
\bauthor{\bsnm{Becattini}, \binits{F.}},
\bauthor{\bsnm{Seidenari}, \binits{L.}},
\bauthor{\bsnm{Del~Bimbo}, \binits{A.}}:
\bctitle{Mantra: Memory augmented networks for multiple trajectory prediction}.
In: \bbtitle{Proceedings of the IEEE Conference on Computer Vision and Pattern
  Recognition}
(\byear{2020})
\end{bchapter}
\endbibitem

%%% 18
\bibitem{marchetti2020multiple}
\begin{botherref}
\oauthor{\bsnm{Marchetti}, \binits{F.}},
\oauthor{\bsnm{Becattini}, \binits{F.}},
\oauthor{\bsnm{Seidenari}, \binits{L.}},
\oauthor{\bsnm{Del~Bimbo}, \binits{A.}}:
Multiple trajectory prediction of moving agents with memory augmented networks.
IEEE Transactions on Pattern Analysis and Machine Intelligence
(2020)
\end{botherref}
\endbibitem

%%% 19
\bibitem{marchetti2022smemo}
\begin{botherref}
\oauthor{\bsnm{Marchetti}, \binits{F.}},
\oauthor{\bsnm{Becattini}, \binits{F.}},
\oauthor{\bsnm{Seidenari}, \binits{L.}},
\oauthor{\bsnm{Del~Bimbo}, \binits{A.}}:
Smemo: Social memory for trajectory forecasting.
arXiv preprint arXiv:2203.12446
(2022)
\end{botherref}
\endbibitem

%%% 20
\bibitem{han2017learning}
\begin{bchapter}
\bauthor{\bsnm{Han}, \binits{X.}},
\bauthor{\bsnm{Wu}, \binits{Z.}},
\bauthor{\bsnm{Jiang}, \binits{Y.-G.}},
\bauthor{\bsnm{Davis}, \binits{L.S.}}:
\bctitle{Learning fashion compatibility with bidirectional lstms}.
In: \bbtitle{Proceedings of the 25th ACM International Conference on
  Multimedia},
pp. \bfpage{1078}--\blpage{1086}
(\byear{2017})
\end{bchapter}
\endbibitem

%%% 21
\bibitem{cui2019dressing}
\begin{bchapter}
\bauthor{\bsnm{Cui}, \binits{Z.}},
\bauthor{\bsnm{Li}, \binits{Z.}},
\bauthor{\bsnm{Wu}, \binits{S.}},
\bauthor{\bsnm{Zhang}, \binits{X.}},
\bauthor{\bsnm{Wang}, \binits{L.}}:
\bctitle{Dressing as a whole: {{Outfit}} compatibility learning based on
  node-wise graph neural networks}.
In: \bbtitle{The {{Web Conference}} 2019 - {{Proceedings}} of the {{World Wide
  Web Conference}}, {{WWW}} 2019},
pp. \bfpage{307}--\blpage{317}
(\byear{2019}).
\doiurl{10.1145/3308558.3313444}
\end{bchapter}
\endbibitem

%%% 22
\bibitem{cucurull2019context}
\begin{bchapter}
\bauthor{\bsnm{Cucurull}, \binits{G.}},
\bauthor{\bsnm{Taslakian}, \binits{P.}},
\bauthor{\bsnm{Vazquez}, \binits{D.}}:
\bctitle{Context-aware visual compatibility prediction}.
In: \bbtitle{Proceedings of the IEEE/CVF Conference on Computer Vision and
  Pattern Recognition},
pp. \bfpage{12617}--\blpage{12626}
(\byear{2019})
\end{bchapter}
\endbibitem

%%% 23
\bibitem{yang2020learning}
\begin{barticle}
\bauthor{\bsnm{Yang}, \binits{X.}},
\bauthor{\bsnm{Du}, \binits{X.}},
\bauthor{\bsnm{Wang}, \binits{M.}}:
\batitle{Learning to {{Match}} on {{Graph}} for {{Fashion Compatibility
  Modeling}}}.
\bjtitle{Proceedings of the AAAI Conference on Artificial Intelligence}
\bvolume{34}(\bissue{01}),
\bfpage{287}--\blpage{294}
(\byear{2020}).
\doiurl{10.1609/aaai.v34i01.5362}
\end{barticle}
\endbibitem

%%% 24
\bibitem{wang2019outfit}
\begin{bchapter}
\bauthor{\bsnm{Wang}, \binits{X.}},
\bauthor{\bsnm{Wu}, \binits{B.}},
\bauthor{\bsnm{Zhong}, \binits{Y.}}:
\bctitle{Outfit compatibility prediction and diagnosis with multi-layered
  comparison network}.
In: \bbtitle{Proceedings of the 27th ACM International Conference on
  Multimedia},
pp. \bfpage{329}--\blpage{337}
(\byear{2019})
\end{bchapter}
\endbibitem

%%% 25
\bibitem{feng2018interpretable}
\begin{botherref}
\oauthor{\bsnm{Feng}, \binits{Z.}},
\oauthor{\bsnm{Yu}, \binits{Z.}},
\oauthor{\bsnm{Yang}, \binits{Y.}},
\oauthor{\bsnm{Jing}, \binits{Y.}},
\oauthor{\bsnm{Jiang}, \binits{J.}},
\oauthor{\bsnm{Song}, \binits{M.}}:
Interpretable partitioned embedding for customized fashion outfit composition.
arXiv preprint arXiv:1806.04845
(2018)
\end{botherref}
\endbibitem

%%% 26
\bibitem{bigi2020automatic}
\begin{bchapter}
\bauthor{\bsnm{Bigi}, \binits{W.}},
\bauthor{\bsnm{Baecchi}, \binits{C.}},
\bauthor{\bsnm{Del~Bimbo}, \binits{A.}}:
\bctitle{Automatic interest recognition from posture and behaviour}.
In: \bbtitle{Proceedings of the 28th ACM International Conference on
  Multimedia},
pp. \bfpage{2472}--\blpage{2480}
(\byear{2020})
\end{bchapter}
\endbibitem

%%% 27
\bibitem{becattini2021plm}
\begin{bchapter}
\bauthor{\bsnm{Becattini}, \binits{F.}},
\bauthor{\bsnm{Song}, \binits{X.}},
\bauthor{\bsnm{Baecchi}, \binits{C.}},
\bauthor{\bsnm{Fang}, \binits{S.-T.}},
\bauthor{\bsnm{Ferrari}, \binits{C.}},
\bauthor{\bsnm{Nie}, \binits{L.}},
\bauthor{\bsnm{Del~Bimbo}, \binits{A.}}:
\bctitle{Plm-ipe: A pixel-landmark mutual enhanced framework for implicit
  preference estimation}.
In: \bbtitle{ACM Multimedia Asia},
pp. \bfpage{1}--\blpage{5}
(\byear{2021})
\end{bchapter}
\endbibitem

%%% 28
\bibitem{he2016deep}
\begin{bchapter}
\bauthor{\bsnm{He}, \binits{K.}},
\bauthor{\bsnm{Zhang}, \binits{X.}},
\bauthor{\bsnm{Ren}, \binits{S.}},
\bauthor{\bsnm{Sun}, \binits{J.}}:
\bctitle{Deep residual learning for image recognition}.
In: \bbtitle{Proceedings of the IEEE Conference on Computer Vision and Pattern
  Recognition},
pp. \bfpage{770}--\blpage{778}
(\byear{2016})
\end{bchapter}
\endbibitem

%%% 29
\bibitem{ge2019deepfashion2}
\begin{bchapter}
\bauthor{\bsnm{Ge}, \binits{Y.}},
\bauthor{\bsnm{Zhang}, \binits{R.}},
\bauthor{\bsnm{Wang}, \binits{X.}},
\bauthor{\bsnm{Tang}, \binits{X.}},
\bauthor{\bsnm{Luo}, \binits{P.}}:
\bctitle{Deepfashion2: A versatile benchmark for detection, pose estimation,
  segmentation and re-identification of clothing images}.
In: \bbtitle{Proceedings of the IEEE/CVF Conference on Computer Vision and
  Pattern Recognition},
pp. \bfpage{5337}--\blpage{5345}
(\byear{2019})
\end{bchapter}
\endbibitem

%%% 30
\bibitem{ahmad2016used}
\begin{bchapter}
\bauthor{\bsnm{Ahmad}, \binits{K.}},
\bauthor{\bsnm{Conci}, \binits{N.}},
\bauthor{\bsnm{Boato}, \binits{G.}},
\bauthor{\bsnm{De~Natale}, \binits{F.G.}}:
\bctitle{Used: a large-scale social event detection dataset}.
In: \bbtitle{Proceedings of the 7th International Conference on Multimedia
  Systems},
pp. \bfpage{1}--\blpage{6}
(\byear{2016})
\end{bchapter}
\endbibitem

%%% 31
\bibitem{he2017mask}
\begin{bchapter}
\bauthor{\bsnm{He}, \binits{K.}},
\bauthor{\bsnm{Gkioxari}, \binits{G.}},
\bauthor{\bsnm{Doll{\'a}r}, \binits{P.}},
\bauthor{\bsnm{Girshick}, \binits{R.}}:
\bctitle{Mask r-cnn}.
In: \bbtitle{Proceedings of the IEEE International Conference on Computer
  Vision},
pp. \bfpage{2961}--\blpage{2969}
(\byear{2017})
\end{bchapter}
\endbibitem

%%% 32
\bibitem{cuffaro2016segmentation}
\begin{bchapter}
\bauthor{\bsnm{Cuffaro}, \binits{G.}},
\bauthor{\bsnm{Becattini}, \binits{F.}},
\bauthor{\bsnm{Baecchi}, \binits{C.}},
\bauthor{\bsnm{Seidenari}, \binits{L.}},
\bauthor{\bsnm{Del~Bimbo}, \binits{A.}}:
\bctitle{Segmentation free object discovery in video}.
In: \bbtitle{European Conference on Computer Vision},
pp. \bfpage{25}--\blpage{31}
(\byear{2016}).
\bcomment{Springer}
\end{bchapter}
\endbibitem

\end{thebibliography}


\end{document}